\documentclass{bmvc2k}

\usepackage{booktabs}
\usepackage{tabularx}
\usepackage[dvipsnames]{xcolor}
\usepackage{amsfonts}
\usepackage{amsmath}
\usepackage{pdfpages}
\usepackage{multido}

\usepackage{amsmath}		
\DeclareMathOperator*{\argmax}{arg\,max}
\usepackage{comment}

\usepackage{tcolorbox}

\title{Lost in Translation? \\ Vocabulary Alignment for Source-Free Adaptation in Open-Vocabulary Semantic Segmentation}

\addauthor{Silvio Mazzucco}{silvio.mazzucco@thegoodailab.org}{1,2*}
\addauthor{Carl Persson}{carl.persson@thegoodailab.org}{1,3*}
\addauthor{Mattia Segu}{mattia.segu@thegoodailab.org}{1,2,4}
\addauthor{Pier Luigi Dovesi}{pier@thegoodailab.org}{1,3}
\addauthor{Federico Tombari}{tombari@google.com}{4,5}
\addauthor{Luc Van Gool}{vangool@vision.ee.ethz.ch}{6}
\addauthor{Matteo Poggi}{m.poggi@unibo.it}{1,7}
\addauthor{\vspace{-1cm}}{}{}

\addinstitution{
  The Good AI Lab \\
  {\begingroup
    \hypersetup{hidelinks}
    \href{https://thegoodailab.org}{thegoodailab.org}
  \endgroup}
}

\addinstitution{
 ETH Zurich
}
\addinstitution{
 AMD Silo AI
}
\addinstitution{
 Google 
}
\addinstitution{
 Technical University of Munich 
}
\addinstitution{
 INSAIT, Sofia University, \\ St. Kliment Ohridski 
}
\addinstitution{
 University of Bologna
 \\
\vspace{0.3cm}
\scriptsize{* joint first authorship}
}

\runninghead{Silvio Mazzucco et al.}{Lost in Translation? VocAlign}

\begin{document}

\maketitle
\begin{abstract}
We introduce \textbf{VocAlign}, a novel source-free domain adaptation framework specifically designed for VLMs in open-vocabulary semantic segmentation. Our method adopts a student–teacher paradigm enhanced with a vocabulary alignment strategy, which improves pseudo-label generation by incorporating additional class concepts. To ensure efficiency, we use Low-Rank Adaptation (LoRA) to fine-tune the model, preserving its original capabilities while minimizing computational overhead. In addition, we propose a \textit{Top-K} class selection mechanism for the student model, which significantly reduces memory requirements while further improving adaptation performance. Our approach achieves a notable \textbf{+6.11 mIoU} improvement on the CityScapes dataset and demonstrates superior performance on zero-shot segmentation benchmarks, setting a new standard for source-free adaptation in the open-vocabulary setting. \\
\end{abstract}

\vspace{-0.3cm}
{\hfill  \small Project page: \url{https://thegoodailab.org/blog/vocalign} \hfill }

\section{Introduction}
\label{sec:intro}
Open-vocabulary semantic segmentation aims to assign a class label to every pixel in an image, extending traditional semantic segmentation by removing the constraint of a fixed set of classes. This is enabled by Vision-Language Models (VLMs), which leverage multimodal synergies and are pre-trained on web-scale datasets. While this ability allows broader generalization, it also makes VLMs more sensitive to domain shifts, which occur in the more complex joint vision-text space, often resulting in poor performance on unseen datasets with different categories.

Unsupervised domain adaptation (UDA) has been widely studied as a solution to domain shifts \cite{wang2018deep}, allowing models to adapt to new domains without requiring annotations for the target domain. However, most UDA approaches \cite{hoyer2022daformer, hoyer2022hrda, hoyer2023mic, araslanov2021self} assume access to the source data on which the original model was trained. This assumption is impractical for VLMs, which are typically trained on proprietary, web-scale datasets that are not publicly available.

\begin{figure}
    \hspace{\columnsep}
    \centering
    \begin{tabular}{cc}
    \includegraphics[height=0.32\columnwidth]{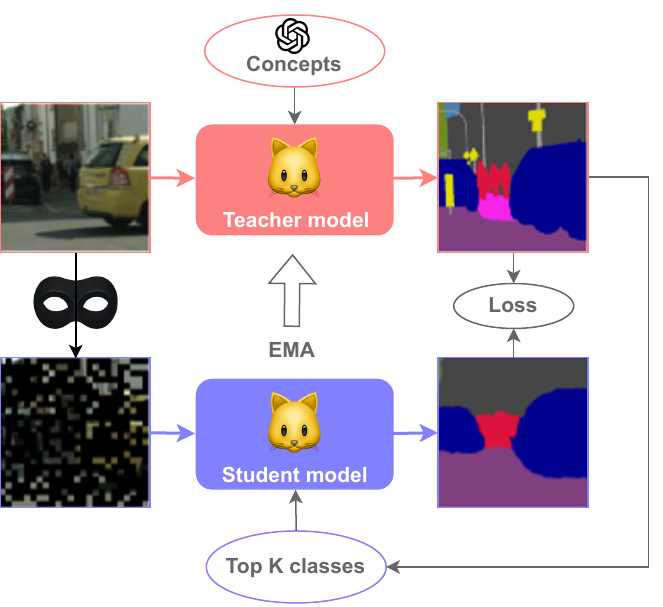} & 
    \includegraphics[height=0.28\columnwidth]{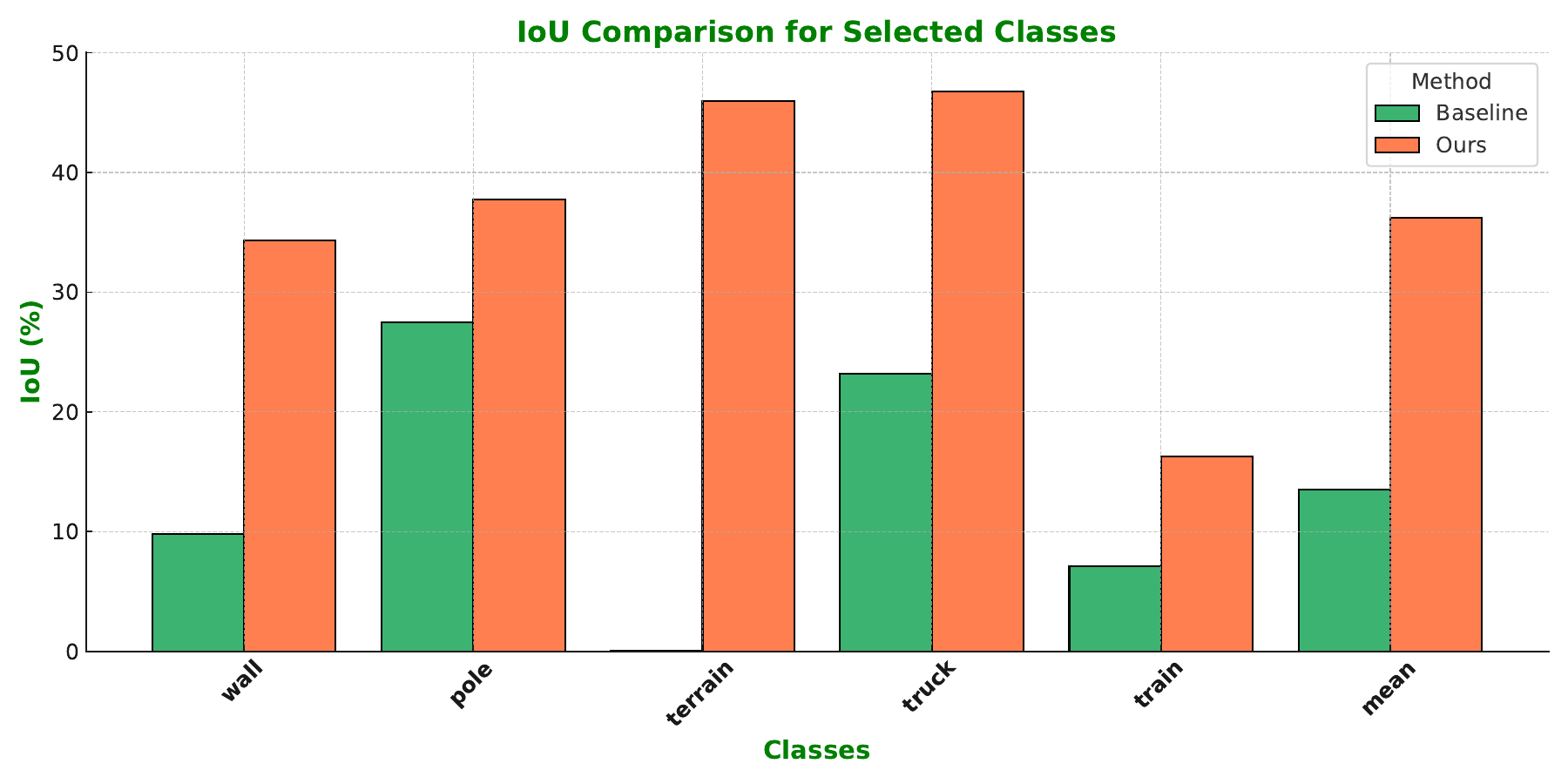} \\
    \end{tabular}    
\caption[VocAlign overview]{\textbf{Method overview.} Left: Our student-teacher framework and the additional techniques it introduces. Right: Impact of VocAlign on selected classes from CityScapes.}
\label{model_overview}
\end{figure}

Source-free domain adaptation (SFDA) addresses this limitation by enabling models to adapt without requiring access to source data. While SFDA offers a more efficient and scalable paradigm, it also introduces challenges due to the absence of annotated samples during adaptation. This approach is particularly well-suited for adapting VLMs, given the impracticality of accessing their massive training datasets. However, existing SFDA methods are primarily designed for models without open-vocabulary capabilities or multimodal vision-language interactions, limiting their applicability to VLMs. Moreover, the sheer size and diversity of VLM training datasets exacerbate adaptation challenges, as distribution shifts and overlapping labels in the feature space become more pronounced.

In this work, we propose \textbf{VocAlign}, the first SFDA framework specifically designed for VLMs in open-vocabulary semantic segmentation. To tackle the unique challenges posed by VLMs, we adopt a student-teacher framework and extend it with techniques tailored to their multimodal and open-vocabulary nature, as shown in Figure \ref{model_overview} (left). First, we enhance pseudo-label generation by augmenting the teacher model's vocabulary with additional class concepts. This leverages the open-vocabulary capabilities of VLMs, improving the alignment between visual embeddings and pseudo-labels across domains, thereby enabling more effective adaptation. Second, to manage the computational cost of adapting large VLMs, we employ parameter-efficient fine-tuning using Low-Rank Adaptation (LoRA) modules \cite{hu2021lora}. This approach preserves the original knowledge learned from web-scale data while minimizing overhead. Third, we introduce a \textit{Top-K} class selection mechanism for the student model, optimizing only a subset of classes per iteration based on pseudo-label predictions. This significantly reduces memory requirements without sacrificing performance.

We evaluate our method on CAT-Seg \cite{cho2023cat}, a state-of-the-art open-vocabulary segmentation model, using the CityScapes dataset. Our approach achieves substantial improvements across several classes, including complete recovery for previously unrecognizable ones, as shown in Figure \ref{model_overview} (right). Furthermore, we extend our evaluation to zero-shot segmentation datasets, demonstrating the effectiveness of our method in diverse settings with varying numbers of target classes. 

Our main contributions are summarized as follows:
\begin{itemize}
    \item We propose the first SFDA framework specifically tailored for open-vocabulary semantic segmentation models built on VLMs.
    \item We introduce a vocabulary-alignment strategy that improves pseudo-label quality by leveraging the multimodal capabilities of VLMs.
    \item We reduce computational complexity through the combination of LoRA modules and \textit{Top-K} class selection, achieving memory efficiency while further improving performance.  
\end{itemize}

\section{Related Work}
\label{sec:related}

We briefly review the literature relevant to our work, covering open-vocabulary semantic segmentation, unsupervised domain adaptation and parameter-efficient fine-tuning approaches.

\textbf{Semantic Segmentation with VLMs.} Semantic segmentation is one of the key tasks \cite{gu2021open, wang2022clip, khan2023introducing} that has greatly benefited from the advent of vision-language models (VLMs) such as CLIP \cite{radford2021learning} and its variants \cite{naeem2025silc, zhou2023non}. Zhou et al. \cite{zhou2022extract} showed that, beyond extracting global feature representations from CLIP, it is possible to directly obtain low-resolution segmentation maps, while Wu et al. \cite{wu2023clipself} refined the extraction and processing pipeline.

A common objective of these methods is to leverage CLIP’s knowledge to handle unseen classes. Li et al. \cite{li2022language} introduced a framework in which text embeddings are explicitly aligned with visual representations, maximizing the correlation between related textual and visual concepts. Similarly, OVSeg \cite{liang2023open} adapts the CLIP backbone to improve the performance of the mask proposal mechanism. Recent work, such as \cite{zhou2023zegclip, xu2023side}, simplifies the pipeline by moving to one-stage methods. Cho et al. \cite{cho2023cat} propose aggregating cost volumes generated by CLIP along both spatial and class dimensions, while Xie et al. \cite{xie2024sed} extend this idea to further enhance accuracy.

\textbf{Unsupervised Domain Adaptation.} Various UDA methods have been developed for downstream tasks such as image classification \cite{long2015learning, ganin2016domain, sun2016deep, long2018conditional}, object detection \cite{chen2018domain, saito2019strong, chen2021scale, li2022cross}, and semantic segmentation \cite{hoffman2016fcns, zou2018unsupervised, chen2019domain, wang2021domain}. Traditional UDA approaches \cite{chen2023pipa, xie2023sepico, hoyer2022daformer, hoyer2022hrda, hoyer2023mic, yang2025micdrop} often assume access to source data during adaptation. However, this assumption is increasingly impractical due to privacy and accessibility concerns. As a result, UDA research has expanded into source-free and test-time adaptation settings, which explicitly address the absence of source data during adaptation.

In test-time adaptation (e.g., \cite{wang2021tent}), the additional challenge is adapting the model online, accounting for the temporal evolution of distribution shifts in the data. Liu et al. \cite{liu2021source} employ adversarial training by generating synthetic samples resembling source-domain data to guide adaptation. Kundu et al. \cite{kundu2021generalize} address domain generalization by leveraging source data during training before applying source-free adaptation to the target domain. Wang et al. \cite{wang2022continual} propose a student-teacher framework with augmentation-averaged pseudo-labels.

More recently, methods tailored to vision-language models have appeared. Samadh et al. \cite{samadh2023align} leverage the alignment of learnable prompts from visual and text encoders during adaptation. Choe et al. \cite{choe2024open} explore open-set domain adaptation for semantic segmentation, although not in a source-free setting.

\textbf{Parameter-Efficient Fine-Tuning Methods.} These methods address the challenge of adapting models with billions of parameters by minimizing the number of learnable parameters while still leveraging the full potential of pre-trained weights.

One of the most popular approaches involves low-rank transformations. Inspired by \cite{hu2021lora}, these methods reparameterize the update delta weights by decomposing them into two lower-rank matrices. Building on this foundation, methods such as \cite{zhang2023adaptive} explore dynamically adjusting the rank of LoRA matrices, while \cite{huang2023lorahub} investigates combining multiple LoRA modules to enhance flexibility and performance.

As an alternative, adapter-based methods have been widely adopted for parameter-efficient fine-tuning \cite{houlsby2019parameter,pfeiffer2020adapterfusion}, while prompt-based tuning \cite{lester2021power,li2021prefix} turns prompts into learnable parameters to improve the accuracy of VLMs with minimal overhead \cite{gao2024clip, zhou2022learning, zhou2022conditional,chowdhury2023apollo}. These parameter-efficient strategies are often applied in the context of UDA \cite{gao2022visual,gan2023decorate}.

\begin{figure*}[t]
    \centering
    \includegraphics[clip,trim=0cm 9.5cm 3cm 0cm,width=\textwidth]{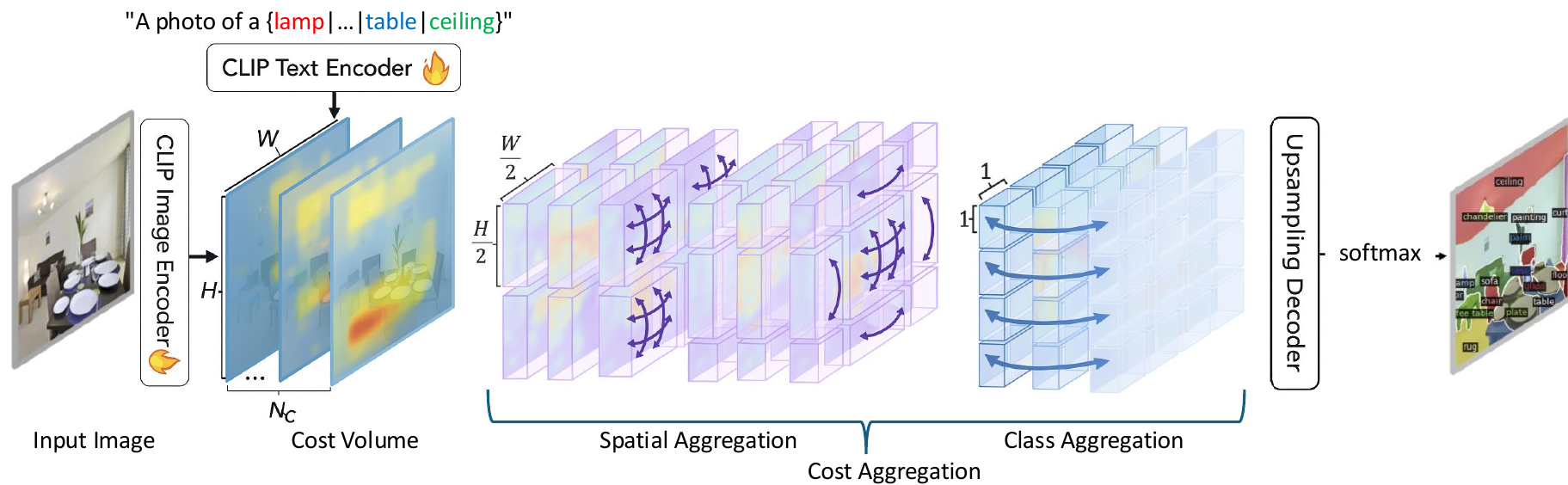} 
    \caption{\textbf{CAT-Seg backbone.} 
    At its core, the cost aggregation module acts at both the spatial and the class levels. Finally, predicted classes are selected through a softmax layer.
    }
    \label{fig:catseg}
\end{figure*}

\section{Method}

We now present our framework, VocAlign. First, we introduce some background about open-vocabulary semantic segmentation and the baseline model, then we describe our teacher-student framework and the specific techniques tailoring it to the open-vocabulary setting.

\subsection{Preliminaries: CAT-Seg Backbone}

VocAlign is applied on CAT-Seg \cite{cho2023cat}, a state-of-the-art model for open-vocabulary segmentation shown in Figure \ref{fig:catseg}, which consists of three components: the feature extractors (CLIP encoders with an additional visual encoder), the cost aggregation module, and the upsampling decoder.

The feature extractors include a slightly modified version of the CLIP image encoder alongside the standard CLIP text encoder. Through this backbone, CAT-Seg extracts textual and dense visual features $\mathcal{D}^L = \phi^L(t)$, $\mathcal{D}^V = \phi^V(x^T)$, which are used to build a cost volume $\mathcal{C} \in \mathbb{R}^{H \times W \times P \times N_c}$ composed of cosine similarities \cite{rocco2017convolutional} between multimodal features:

\small\begin{equation} \mathcal{C}(i,n) = \frac{\mathcal{D}^V(i) \cdot \mathcal{D}^L(n)}{\parallel \mathcal{D}^V(i) \parallel \parallel \mathcal{D}^L(n) \parallel}, \end{equation}\normalsize

\noindent where $i$ denotes pixel coordinates and $n$ corresponds to the text embedding of one of the $N_c$ classes. Text embeddings are enriched with $P$ diverse prompts, such as ``a painting of a \textit{class}'' or ``a rendering of a \textit{class},'' resulting in embeddings of shape $\phi^L \in \mathbb{R}^{N_c \times P \times d_L}$.

To refine the coarse cost volume, CAT-Seg employs a cost aggregation module with two distinct aggregation mechanisms, both implemented with Swin Transformers:  
i) \textit{Spatial Aggregation}, modeling pixel-level spatial interactions to propagate information; and  
ii) \textit{Class Aggregation}, focusing on interactions among different semantic classes.  

After refinement, the decoder upsamples the volume to match the input resolution. The output is then passed through a softmax layer to produce the final predictions.

\subsection{Student-teacher framework}

As the baseline for VocAlign, we adopt student-teacher knowledge distillation \cite{tarvainen2017mean} to transfer knowledge from the source to the target domain. Initially, a neural network $f_{\theta}$ is trained on the source domain using source images $\mathcal{X}^S = \{x^s_k\}^{N_s}_{k=1}$ and their respective labels $\mathcal{Y}^S = \{y^s_k\}^{N_s}_{k=1}$. In the context of semantic segmentation, the most common loss is the pixel-wise cross-entropy:

\small\begin{equation}
\mathcal{L}^{S,seg}_k = \mathcal{H}(f_\theta(x^S_k), y^S_k), \quad\quad \text{with} \quad\quad \mathcal{H}(\hat{y}, y) = -\sum^H_{i=1}\sum^W_{j=1}\sum^C_{c=1}y_{ijc}\log{\hat{y}_{ijc}}
\end{equation}\normalsize

\noindent In the SFDA setting we target, source images and labels are no longer available after deployment. From this point onward, training data refers to target images $\mathcal{X}^T = \{x^T_k\}^{N_T}_{k=1}$, which lack labels.

The pre-trained model serves as the \textit{teacher} $g_\phi$ in our SFDA framework, generating pseudo-labels that supervise a second instance of the model, the \textit{student} $f_\theta$, which is adapted to the target domain. Pseudo-labels are generated by feeding target images to $g_\phi$:

\small\begin{equation}
p_{i,j}^{T} = [c = \argmax_{c'}g_\phi(x^{T})_{i,j,c'}],
\end{equation}\normalsize

\noindent where $[\cdot]$ denotes the Iverson bracket. We also weight the loss by exploiting class probabilities predicted by $g_\phi$, computing the ratio of pixels whose confidence exceeds a threshold $\tau$. The confidence $q^T$ is represented by the pixelwise maximum softmax probability:

\small\begin{equation}
q_{i,j}^{T} = \frac{[\sum^H_{i=1}\sum^W_{j=1}\max_{c'}g_\phi(x^{T})_{i,j,c'} > \tau]}{H \cdot W} 
\end{equation}\normalsize

During adaptation, the teacher remains frozen with no gradients backpropagated through it. However, it is updated over time as an exponential moving average (EMA) of the student weights \cite{tarvainen2017mean}, specifically:  
$\phi_{t+1} \leftarrow \alpha\phi_{t} + (1 - \alpha)\theta_{t}$,  
where $\alpha \in [0,1]$ controls how much the student update influences the teacher.

\textbf{Student input masking.}  
Following \cite{hoyer2023mic}, we apply masking to the images fed to $f_\theta$ to improve adaptation. A binary mask is sampled as:

\small\begin{equation}
\mathcal{M}_{\substack{mb+1:(m+1)b \\ nb+1:(n+1)b}} = [v > r] \quad \text{with} \quad v \sim \mathcal{U}(0,1),
\end{equation}\normalsize

\noindent where $[\cdot]$ is the Iverson bracket, $b$ the patch size, $r$ the mask ratio, $m \in [0, ..., W/b - 1]$, $n \in [0, ..., W/b - 1]$ the patch indices, and $\mathcal{U}(0,1)$ a uniform distribution. Student predictions $\hat{y}^M$ are thus based only on the visible context of the masked image:

\small\begin{equation}
\hat{y}^M = f_\theta(x^M) \quad\quad \text{with} \quad\quad x^M = \mathcal{M} \odot x^T
\end{equation}\normalsize

Finally, the student is supervised by a cross-entropy loss $\mathcal{H}$ between its predictions and the teacher’s pseudo-labels:  
$\mathcal{L}^M = q^T\mathcal{H}(\hat{y}^M, p^T)$,  
where $p^T$ are pseudo-labels and $q^T$ is the confidence-based weight.

\subsection{Vocabulary Alignment}
\begin{figure}[t]
    \hspace{\columnsep}
    \centering
    \begin{tabular}{cc}
        \includegraphics[clip,trim=0cm 0.7cm 0cm 0.5cm,width=0.5\columnwidth]{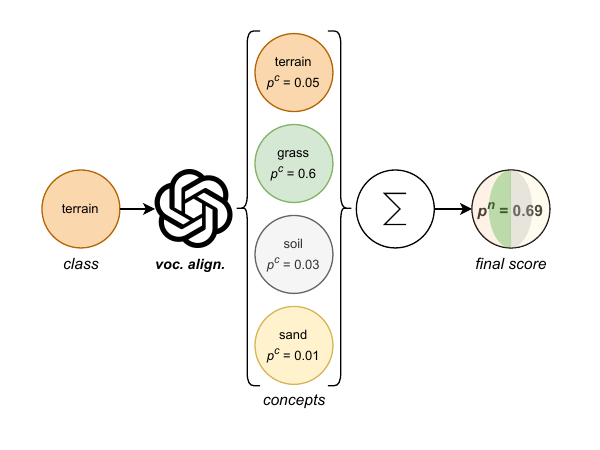} & 
        \includegraphics[clip,trim=0cm 1cm 0cm 1cm,width=0.5\columnwidth]{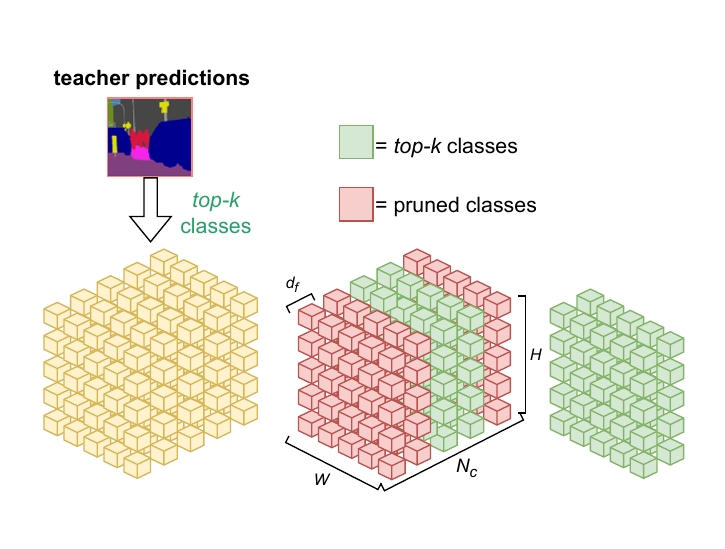} \\
    \end{tabular}     
    \caption{\textbf{VocAlign core components.} Left: example of Vocabulary Alignment on CityScapes. Right: Top-K selection in action.}
    \label{fig:voc_align}
\end{figure}

To address the challenges of the open-vocabulary setting, VocAlign introduces data augmentation in the text space, rather than relying solely on image augmentations \cite{wang2022continual}. We enrich target dataset classes with \textit{concepts}—additional text descriptions or synonyms, used only during adaptation. Since the source model is pre-trained on massive datasets, it is biased toward classes present in its training data, which may overlap only partially or under different names with target classes. This is a new issue in SFDA for VLMs, as domain shifts involve not only visual appearance but also vision-text alignment.

For example, if the class ``animal'' was present during pre-training, adding it as a concept to the target class ``cat'' can improve performance, especially when no other animal-related classes exist in the target dataset. Accordingly, the cost volume is expanded to $n_{tot}$ classes, including the original ones plus concepts. Each class $n$, $n = 1, \ldots, N_c$, can be augmented with $c(n)$ concepts. Before pseudo-label generation, concepts are aggregated back to their original class:  

\small\begin{equation}
p_{ij}^c = \frac{e^{x_{ij}^c}}{\sum_{c'=1}^{n_{\text{tot}}} e^{x_{ij}^{c'}}}, \quad c = 1, \ldots, n_{tot}, \quad\quad 
p_{ij}^{n} = \sum^{C(n)}_{h=1}p_{ij}^{c_h(n)}, \quad n = 1, \ldots, N_c
\end{equation}\normalsize

\noindent where $p^c_{ij}$ is the probability over $n_{tot}$ classes including concepts, and $p^n_{ij}$ is the final probability for the $N_c$ original classes. An example is shown in Figure \ref{fig:voc_align} (left).

To streamline concept creation for large datasets, we generate them automatically using ChatGPT o1, with prompts designed to capture context, objectives, and dataset information. More details are reported in the supplementary material.

\subsection{Top-K selection}

Adapting to many classes can be computationally prohibitive. To address this, we use teacher predictions (enhanced by concepts) to identify likely classes in an image. Given teacher probabilities $\hat{y} \in \mathbb{R}^{N_c, H, W}$, we compute the mean activation per class and select the \textit{Top-K} with the highest averages. The remaining classes are pruned from the cost volume before aggregation, as illustrated in Figure \ref{fig:voc_align} (right).

As a result, only a subset of classes receives supervision per iteration. Although this may seem to reduce supervision, the higher-quality pseudo-labels produced by the teacher ensure that present classes are reliably predicted, ultimately providing stronger supervision for those classes.

\section{Experiments}

We now introduce our experimental results. First, we describe the implementation details, the datasets used in our evaluation, and the training scheduling, then we report our main experiments and conclude with some analysis and ablation studies.

\subsection{Implementation Details}
Our method is based on the \textit{mmsegmentation} framework \cite{contributors2020mmsegmentation} and its CAT-Seg implementation, and we build our student-teacher framework integrating the code provided by \cite{hoyer2023mic}. 
We used CAT-Seg backbone with Resnet-101 and ViT-B encoders, modified to use 10 prompt templates repeated eight times, rather than the 80 templates in the original model, for efficiency -- we will also provide results for the full 80 templates.
We start from CAT-Seg weights being pre-trained on COCO-Stuff \cite{caesar2018coco}, and we adapt these weights in any of the experiments we report about. 
We introduce LoRA exclusively in the CLIP backbone, to the first 4 layers of the ViT for the image encoder and of the transformer for the text encoder, in the attention projection matrices. In the standard setting, we use a LoRA rank equal to 2. We keep the rest of the model frozen.

\textbf{Datasets.} We mainly use CityScapes for training and evaluation, as it is a commonly utilized dataset to evaluate domain adaptation methods, and because it is sufficiently different from COCO-Stuff to serve as a favorable target domain. It is comprised of 2975 training images and 500 validation samples, with a total of 19 classes. To proof that our method also generalizes to other datasets, we use ADE20K-150 \cite{zhou2019semantic} and PASCAL-Context 59 \cite{mottaghi2014role}. These datasets were utilized to measure the zero-shot performance of the original CAT-Seg model. ADE20k-150 comprises 20k training and 2k validation images, annotated with a total of 150 classes. PASCAL-Context 59 contains 5k training and validation images, with 59 class labels. The mean Intersection over Union (mIoU) is used as the main metric.

\textbf{CityScapes Training.} We train our method for 40k iterations with the AdamW optimizer \cite{loshchilov2017decoupled}, a learning rate of $5 \times 10^{-5}$ after a quick warm-up phase at the beginning of the training, starting at $5 \times 10^{-6}$ for 500 iterations. We use a batch size of 2 and a Top-K selection value of 15. The input data is augmented by randomly cropping the images with $512 \times 512$ crops, and by applying random color jitter. The image is masked with a masking ratio of 0.7. Similarly to the majority of the methods, we set $\alpha =0.99$ for the EMA update of the teacher. Suitable concepts were chosen manually based on the CityScapes class descriptions.

\textbf{Multi-dataset Training.} To benchmark the generalizability of our method we choose a single configuration to apply to all three datasets. We train for 15k iterations and use a batch size of 2 when it is applicable, otherwise, a batch size of 1 is applied. The input images are augmented by random cropping images to $512 \times 512$ crops and by applying random flipping, photometric distortion, and color jitter. The learning rate is chosen to be $3 \times 10^{-5}$ and we apply a masking ration of 0.5. Concepts are generated using ChatGPT-o1 for all datasets. We utilize a Top-K value of 50 when training the modified 10 prompt template CAT-Seg model and a Top-K value of 35 when training on the full CAT-Seg model.

\begin{table*}[t]
    \centering
    \setlength{\tabcolsep}{6pt}
    \resizebox{\columnwidth}{!}{
    \begin{tabular}{l|ccccccccc|c}
        \toprule
        Method & car & fence & road & mbike & ... & pole & truck & wall & terrain
        & mIoU \\
        \midrule
        \midrule
        Zero-Shot CAT-Seg & \bf 76.38 & \bf 38.37 & \bf 86.03         
        & \bf 55.35 & ... & 27.47 & 23.22 & 9.77 & 0.02
        & 47.56 \\
        \bf VocAlign (ours) & 67.48 & 31.70 & 84.87 & 54.83 & ... & 37.71 & \bf 46.76 & \bf34.32 & \bf 45.95 & \bf 53.67 \\
        \textcolor{darkgray}{Improve (\%)} & \textcolor{RedOrange}{-8.9} & \textcolor{RedOrange}{-6.67} & \textcolor{RedOrange}{-1.16} & \textcolor{RedOrange}{-0.52} & ... & \textcolor{ForestGreen}{+10.24} & \textcolor{ForestGreen}{+23.54} & \textcolor{ForestGreen}{+24.55} & \textcolor{ForestGreen}{+45.93} & \textcolor{ForestGreen}{+6.11} \\
        \bottomrule
    \end{tabular}}
    \caption{\textbf{Per-class results for top 4 best and worst classes on CityScapes.} We compare VocAlign to the Zero-Shot predictions by CAT-Seg. Last column shows mIoU on all classes.}
    \label{tab:cityscapes}
\end{table*}

\subsection{Main Results}

\textbf{Results on CityScapes.} Our initial results on CityScapes are shown in Table \ref{tab:cityscapes}, for top-4 best and worst classes improved by VocAlign (complete results in the supplementary material). VocAlign demonstrates a strong capability to adapt, especially for classes that perform poorly in the zero-shot setting. This includes the \textit{terrain} class, which improves drastically from a nigh zero mIoU evaluation. Many of the large improvements are thought to stem from ambiguities in the class names and their descriptions. For example, the \textit{wall} class in CityScapes is defined as an \textit{individual standing wall, not part of a building}. However, since \textit{wall} is the only word delivered to the segmentation model, this additional class information is lost. This is something we can recover by adding concepts or modifying the teacher classes, which is likely the cause for the large improvements in this particular class. Furthermore, we found that our method decreases the performance for certain classes -- most notably cars. This is thought to be caused by the hood of the capturing vehicle, which is not masked during training. The model therefore learns to classify the general area of the hood and, to some degree, the nearby road, as a car. The hood is then masked during validation, which results in parts of the road being misclassified as a car, reducing the score of this particular category, as shown in Figure \ref{fig:visualization}. We further confirm this hypothesis in the supplementary material.

\begin{figure*}[t]
    \centering
    \includegraphics[width=\textwidth]{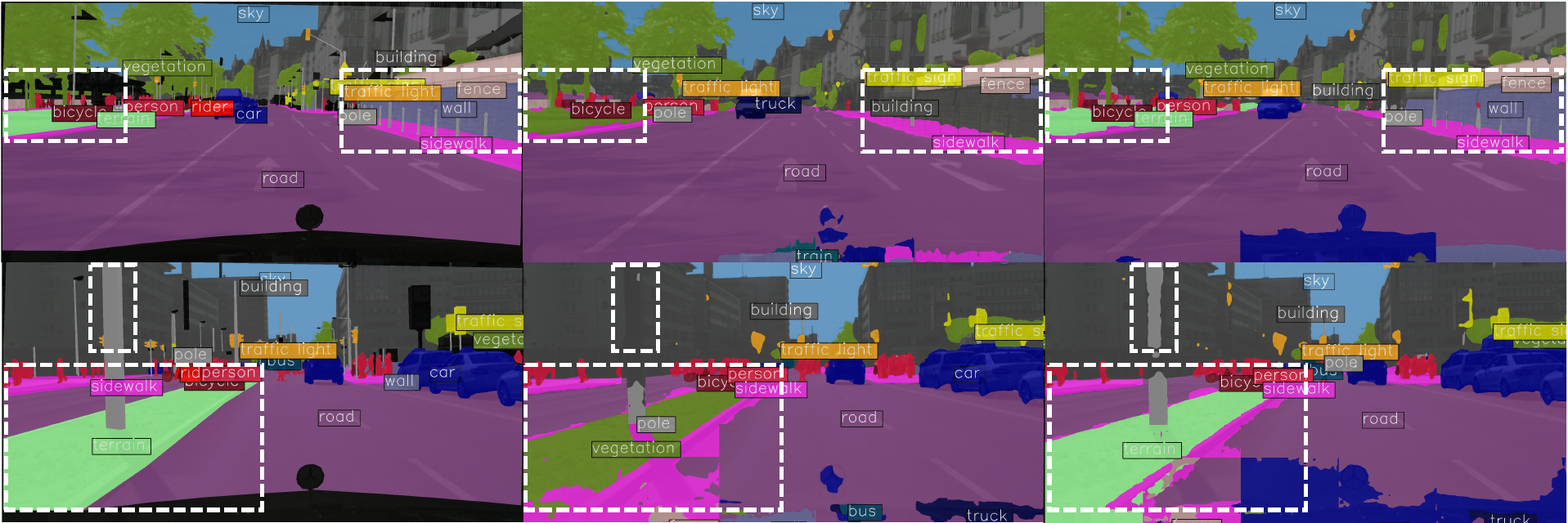} 
    \caption{\textbf{Qualitative results on CityScapes dataset.} 
   From left to right: ground truth segmentation map, the zero-shot CAT-Seg prediction, and VocAlign prediction.}
    \label{fig:visualization}
\end{figure*}

\textbf{Multi-dataset results.} Our evaluation on multiple datasets is shown in Table \ref{tab:prompt10}. Accordingly, our method can generalize well to other datasets using a generic configuration, where the model adaptability remains strong. The decrease in performance, compared to the CityScapes training, can be partly explained by the use of prompts generated by ChatGPT and the use of a lower number of iterations for all datasets. The method was also applied to the full CAT-Seg model (right), where we continue to exhibit moderate adaptability. The reason for the lower evaluation score is likely due to the low Top-K selection value, constrained by our memory restrictions -- a single 64GB A100 GPU.

\textbf{Comparison to other methods.} To our knowledge, there are no prior works covering SFDA on open-vocabulary semantic segmentation models. As such, we implement a simple baseline utilizing the minimization of entropy as the training objective. This objective makes it possible to adapt the CAT-Seg model in a source-free manner. The results of this method can be viewed in Table \ref{tab:prompt10}. The entropy minimization objective can adapt the CAT-Seg model, although only to a limited degree in comparison to VocAlign.

\begin{table}[t]
    \centering
    \setlength{\tabcolsep}{3pt}
    \resizebox{\columnwidth}{!}{
    \begin{tabular}{ccc}
    \begin{tabular}{l|ccc}
        \toprule
        Method & CityScapes & PC-59 & ADE20k-150 \\
        \midrule
        \midrule
        Zero-Shot & 47.56 & 55.52 & 26.88 \\
        Min-Entropy & 47.67 & 55.87 & 27.23 \\
        Teacher-Student & 44.58 & 55.20 & 26.55 \\
        + Masking & 47.71 & 55.89 & 26.84 \\
        + Vocab Alignment & 49.58 & 56.81 & 26.77 \\
        + TopK & \textbf{49.58} & \textbf{57.01} & \textbf{27.39} \\
        \textcolor{darkgray}{Improve (\%)} & \textcolor{ForestGreen}{+2.02} & \textcolor{ForestGreen}{+1.49} & \textcolor{ForestGreen}{+0.51} \\
        \bottomrule
    \end{tabular}
    & \hspace{0.5cm} &
    \begin{tabular}{l|ccc}
        \toprule
        Method & CityScapes & PC-59 & ADE20k-150 \\
        \midrule
        \midrule
        Zero-Shot & 47.88 & 56.94 & 27.22 \\
        Min-Entropy & 47.95 & - & - \\
        \bf VogAlign (ours) & \textbf{48.97} & \textbf{57.32} & \textbf{27.40} \\ 
        \textcolor{darkgray}{Improve (\%)} & \textcolor{ForestGreen}{+1.09} & \textcolor{ForestGreen}{+0.38} & \textcolor{ForestGreen}{+0.18} \\
        \bottomrule
    \end{tabular}
    
    \end{tabular}
    }
    \caption{\textbf{Multi-dataset evaluation.} Left: 10-prompt CAT-Seg, Min-Entropy baseline and ablated versions of VocAlign. Right: original 80-prompt CAT-Seg.
    }
    \label{tab:prompt10}
\end{table}

\begin{figure}[t]
    \centering
    \setlength{\tabcolsep}{1pt}
    \begin{tabular}{cc}
        \includegraphics[width=\columnwidth]{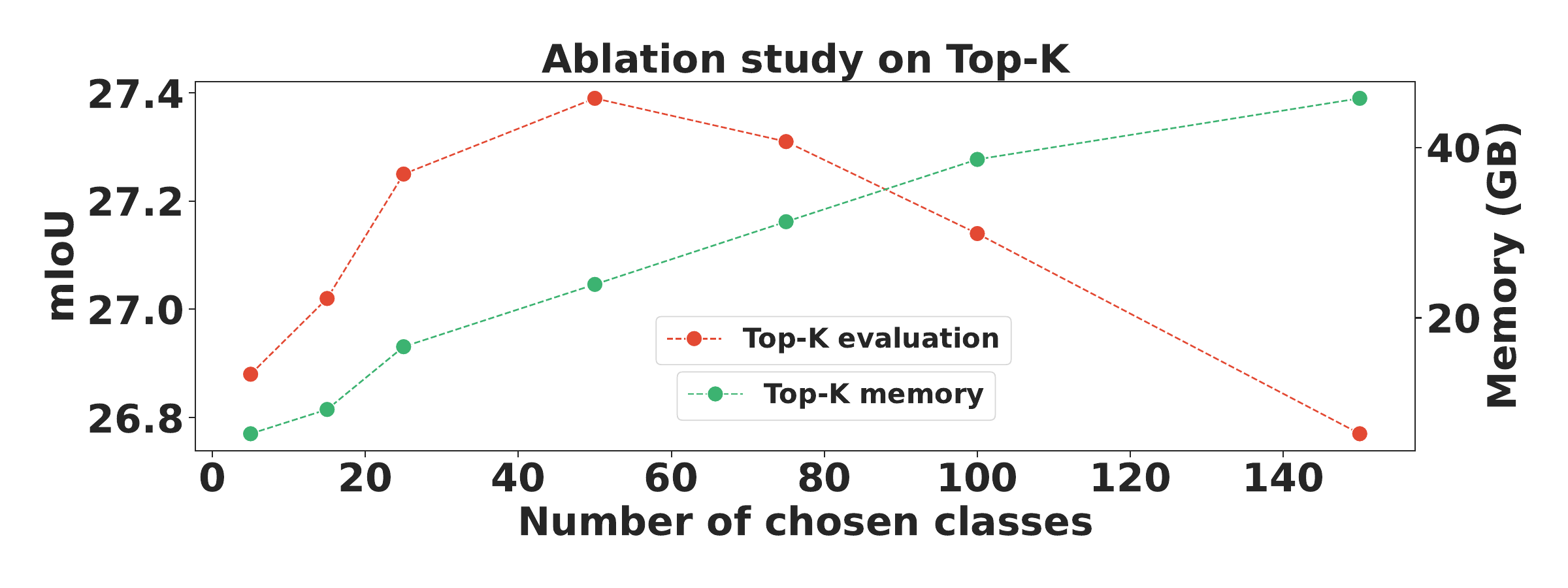} \\
    \end{tabular}
    \caption{\textbf{Ablation study on ADE20k-150 -- impact of the number of classes in Top-K.} We vary the number of classes selected by the Top-K method.}
    \label{fig:ablation_topk}
\end{figure}

\begin{figure}[t]
    \centering
    \setlength{\tabcolsep}{1pt}
    \begin{tabular}{cc}
        \includegraphics[width=\columnwidth]{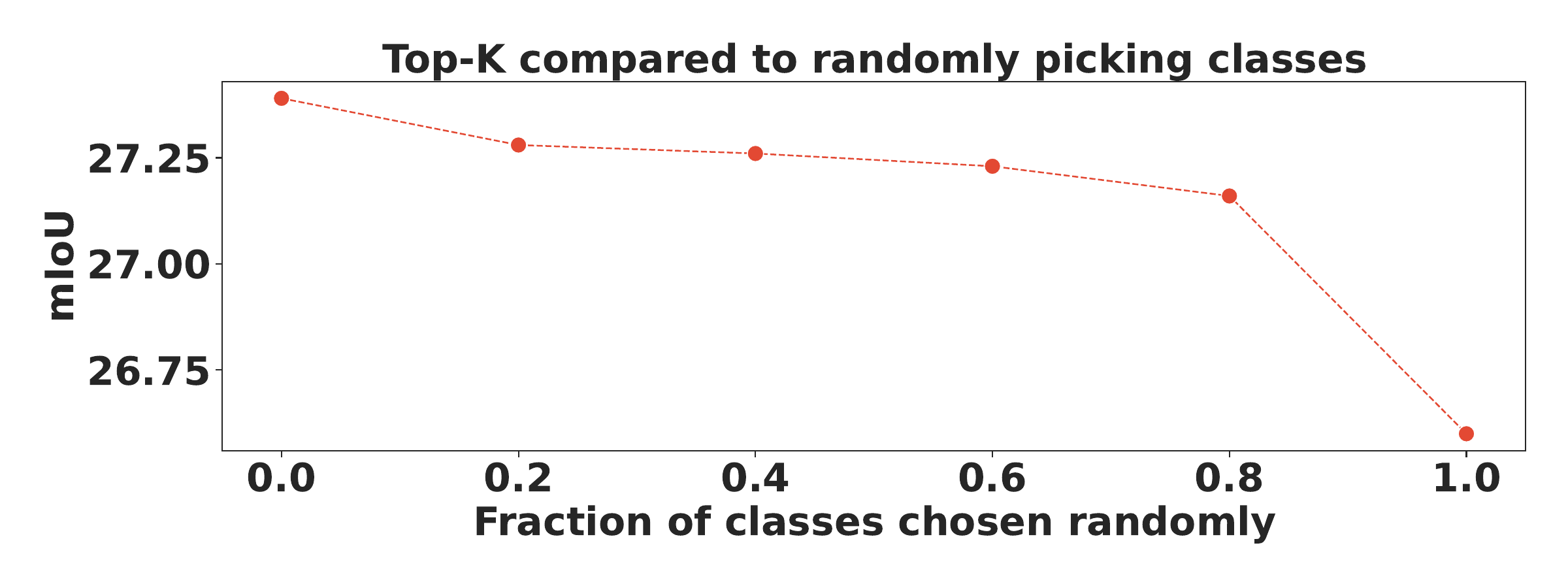} \\
    \end{tabular}
    \caption{\textbf{Ablation study on ADE20k-150 -- Top-K selection vs random selection.} We replace a fraction of Top-K classes with randomly sampled ones.} 
    \label{fig:ablation_topk_random}
\end{figure}

\subsection{Analysis and Ablations}

\textbf{Model component analysis.} Table \ref{tab:prompt10} (left) shows the effectiveness of each model component. The baseline Teacher-Student setup performs poorly and fails to meaningfully adapt the model to any of the datasets. Adding masking to the model significantly improves upon the baseline, yet the increase over CAT-Seg is only moderate or inexistent, depending on the dataset. Our vocabulary alignment method is capable of significantly adapting the model on CityScapes and PASCAL-Context 59, however, we notice a decrease in performance with respect to the use of masking alone on ADE20K-150. This could be caused by the large amount of classes in the dataset, making it hard to find suitable concepts. Finally, incorporating the Top-K method moderately improves the performance of the model on any dataset.

\textbf{Final model performance.} Table \ref{tab:prompt10} (right) demonstrates that our method can be applied to the full CAT-Seg model utilizing all 80 prompt templates. However, due to the higher memory requirements, a lower Top-K value of 35 was chosen to fit a single GPU. This results in a more moderate performance increase. Nevertheless, we still recognize a strong improvement on CityScapes and modest improvements for PC-59 and ADE20k-150 compared to the zero-shot setting.

\textbf{Impact of Top-K selection.} Figures \ref{fig:ablation_topk} and \ref{fig:ablation_topk_random} proves the effects of Top-K selection on ADE20k-150. Top-K was initially implemented as a cost-saving method -- as highlighted by the green curve in Figure \ref{fig:ablation_topk}. However, this ablation also unveils that it results in a moderate performance gain -- see red curve. 
Moreover, we show in Figure \ref{fig:ablation_topk_random} how replacing a portion of the Top-K selected classes with random ones yields worse performance, confirming that Top-K selection is strictly better than picking classes randomly. 
The increase in performance by using Top-K implies that the most relevant classes receive relatively stronger gradients, whereas the pruned classes do not receive wrong supervision.
\section{Conclusions}
We introduced VocAlign, the first-ever SFDA method for open-vocabulary semantic segmentation models. This is done through a vocabulary alignment strategy for pseudo-labels generation, a \textit{Top-K} classes selection mechanism on the student model to decrease memory requirements, and a clever use of LoRAs. Our method shows great results on CityScapes, which we consider our main benchmark since it is one of the most used datasets in UDA. Moreover, we show promising results on other open-vocabulary datasets, with which we demonstrate addressing issues related to the nature of VLMs. This work opens the path for more research on this topic, which could become even more relevant in the future.

\textbf{Acknowledgment.} We acknowledge the European High Performance Computing Joint Undertaking (EuroHPC JU), EuroCC National Competence Center Sweden (ENCCS) and the CINECA award under the ISCRA initiative for the availability of high-performance computing resources and support.

\bibliography{main,morebibs,egbib}
\multido{\i=1+1}{5}{


\section*{Supplementary material}

# Lost in Translation? Vocabulary Alignment for Source-Free Adaptation in Open-Vocabulary Semantic Segmentation – Supplementary Material

Silvio Mazzucco[1,2*]
silvio.mazzucco@thegoodailab.org

Carl Persson[1,3*]
carl.persson@thegoodailab.org

Mattia Segu[1,2,4]
mattia.segu@thegoodailab.org

Pier Luigi Dovesi[1,3]
pier@thegoodailab.org

Federico Tombari[4,5]
tombari@google.com

Luc Van Gool[6]
vangool@vision.ee.ethz.ch

Matteo Poggi[1,7]
m.poggi@unibo.it

[1] The Good AI Lab
thegoodailab.org

[2] ETH Zurich

[3] AMD Silo AI

[4] Google

[5] Technical University of Munich

[6] INSAIT, Sofia University,
St. Kliment Ohridski

[7] University of Bologna

* joint first authorship

## Introduction

This document provides supplementary material for BMVC paper "Lost in Translation? Vocabulary Alignment for Source-Free Adaptation in Open-Vocabulary Semantic Segmentation." The supplementary content elaborates on key aspects of the work, addressing implementation details, detailed analyses, and additional results.

Specifically, this document includes:

- **ChatGPT Prompts and Concept Augmentation:** Insights into the prompt engineering process and examples of how generated concepts enhance model performance.

- **Analysis of Ego-Vehicle Faulty Segmentation:** Examination of common segmentation errors, particularly for ego-vehicle-related regions, and the applied solutions.

- **Class-by-Class Analysis:** Detailed per-class performance evaluation across multiple datasets, highlighting areas of significant improvement.

- **Qualitative Results:** Visual comparisons illustrating the model's improvements in segmentation quality on various datasets.

---

**ChatGPT o1 Prompt**

Context: An open vocabulary semantic segmentation model was trained on a dataset with a set amount of labels. The model is based on CLIP, which makes it possible to extend this model to other datasets with other labels. However, the labels of the new evaluation dataset might not properly align with the labels used for training. The following task aims to solve this problem through the modification of the labels used for evaluation.

Task: Given a list of training labels and a list of evaluation labels, the objective is to find a good mapping of training labels to the evaluation labels. Multiple training labels can be mapped to the same evaluation label, but one training label can not appear more than once.

If the evaluation label does not have an exact match, then map two more synonyms, chosen freely, beyond the already mapped training labels. Also modify the evaluation label to align with the description if necessary.

Evaluation labels (with descriptions):
...

Training labels (with descriptions):
...

---

**Follow-up ChatGPT o1 Prompt**

For all evaluation labels, find a synonym and add it to the list. The synonym does not have to be from any previous list and can be chosen completely freely.

---

Figure 1: **Initial and Follow-up prompts used in ChatGPT o1.**

# 1 ChatGPT Prompts and Concept Augmentation

In this section, we describe how we obtain augmented concepts to enhance teacher performance during adaptation.

## 1.1 Prompt Design

To automatize the generation of augmented concepts, we interact with ChatGPT o1 making it responsible of the concepts generation process. To generate favourable concepts we feed ChatGPT with the following information: the context of the task with its corresponding restrictions, as well as information on the target labels and the source training labels. The task consists of finding a good mapping from training labels to the target labels. If this mapping does not exist, ChatGPT is asked to generate synonyms instead.

We also found that the descriptions of the labels in the evaluation datasets did not align well with the visual semantics of such labels. In this case, we also ask ChatGPT to find a suitable replacement label that aligns better with the label description. In practice, this was seldom the case but helped in certain edge case scenarios. We found that this prompt could generate satisfactory concepts for all datasets. However, for larger datasets, the amounts of synonyms to generate was changed to satisfy computational restrictions, but only slight

| Original Classes | ChatGPT Concepts |
|---|---|
| road | street |
| sidewalk | pavement, floor-stone, floor-tile, footpath |
| building | building-other, house, skyscraper, structure |
| wall | wall-other, solid-other, structural-other, barrier |
| fence | railing, enclosure |
| pole | light, metal, post |
| traffic light | signal |

| Original Classes | ChatGPT Concepts |
|---|---|
| traffic sign | stop sign, parking meter, banner, signboard |
| vegetation | tree, bush, plant-other, foliage |
| terrain | grass, ground-other, dirt, landscape |
| sky | sky-other, heavens |
| person | human |
| rider | horse, skateboard, cyclist |
| car | automobile |

| Original Classes | ChatGPT Concepts |
|---|---|
| truck | lorry |
| bus | coach |
| train | locomotive |
| motorcycle | motorbike |
| bicycle | bike |

Table 1: **ChatGPT Concepts.** List of the original classes and the additional concepts generated by ChatGPT o1 for CityScapes.

| Method | road | swalk | build. | wall | fence | pole | tlight | sign | veg. | terrain | sky | person | rider | car | truck | bus | train | mbike | bike | mIoU |
|---|---|---|---|---|---|---|---|---|---|---|---|---|---|---|---|---|---|---|---|---|
| Zero-Shot | **86.03** | 48.58 | 81.79 | 9.77 | **38.37** | 27.47 | 46.09 | **50.05** | 84.08 | 0.02 | 81.69 | 68.08 | **0.05** | **76.38** | 23.22 | 49.18 | 7.12 | **55.35** | **70.23** | 47.56 |
| Ours | 84.87 | **50.70** | **84.90** | **34.32** | 31.70 | **37.71** | **47.59** | 49.65 | **87.12** | **45.95** | **88.70** | **69.06** | 0.00 | 67.48 | **46.76** | **52.35** | **16.28** | 54.83 | 69.72 | **53.67** |
|  | -1.16 | +2.12 | +3.11 | +24.55 | -6.67 | +10.24 | +1.50 | -0.40 | +3.04 | +45.93 | +7.01 | +0.98 | -0.05 | -8.9 | +23.54 | +3.17 | +9.16 | -0.52 | -0.51 | +6.11 |

Table 2: **Per-class results on CityScapes.** We compare our proposed method to the initial Zero-Shot predictions by CAT-Seg.

variations to the prompt were made.

## 1.2 Examples of Augmented Concepts

By exploiting the prompts introduced before, we can generate augmented concepts related to any of the semantic classes in a dataset. In Table 1, we show some examples of concepts generated by ChatGPT o1 specifically for the original semantic classes in the CityScapes dataset.

# 2 Ego-Vehicle Faulty Segmentation

For the sake of completeness, in Table 2 we report the results concerning all of the single classes on CityScapes. As mentioned in the main paper, we observe a weird worsening of the performance on the class *car* in the CityScapes results. We assume that this mainly

| Method | road | swalk | build. | wall | fence | pole | tlight | sign | veg. | terrain | sky | person | rider | car | truck | bus | train | mbike | bike | mIoU |
|---|---|---|---|---|---|---|---|---|---|---|---|---|---|---|---|---|---|---|---|---|
| Zero-Shot Crop | 89.2 | 55.25 | 82.3 | 13.52 | 38.55 | 27.51 | 46.09 | 51.26 | 84.45 | 0.02 | 86.58 | 68.12 | 0.05 | 81.19 | 24.13 | 49.65 | 13.92 | 55.48 | 70.99 | 49.38 |
| Ours Crop | 88.91 | 57.06 | 85.97 | 35.89 | 31.65 | 38.77 | 48.69 | 54.56 | 87.2 | 47.07 | 90.14 | 69.11 | 0.0 | 85.67 | 47.1 | 53.31 | 17.34 | 55.17 | 70.15 | 55.99 |
| | -0.29 | +1.81 | +3.67 | +22.37 | -6.90 | +11.26 | +2.60 | +3.30 | +2.75 | +47.05 | +3.56 | +0.99 | -0.05 | +4.48 | +22.97 | +3.66 | +3.42 | -0.31 | -0.84 | +6.61 |

**Table 3: Per-class results on CityScapes when cropping.** We compare our proposed method to the initial Zero-Shot predictions by CAT-Seg when cropping the images to get rid of the ego-vehicle during evaluation.

| Method | plane | bag | bed | bedcloth. | bench | bike | bird | boat | book | bottle | build | bus | cabinet | car | cat | ceil. | chair | cloth | computer | mIoU |
|---|---|---|---|---|---|---|---|---|---|---|---|---|---|---|---|---|---|---|---|---|
| Zero-Shot | 88.82 | 43.08 | 25.47 | 40.42 | 26.78 | 76.64 | 90.35 | 76.47 | 52.54 | 79.49 | 42.19 | 94.24 | 42.32 | 74.69 | 91.21 | 52.76 | 56.35 | 14.00 | 14.96 | 55.52 |
| Ours | 89.02 | 41.78 | 25.43 | 42.61 | 27.42 | 77.04 | 90.68 | 78.66 | 56.89 | 79.68 | 39.18 | 94.37 | 43.18 | 79.08 | 91.90 | 50.50 | 47.07 | 19.39 | 18.25 | 57.01 |
| | +0.20 | -1.30 | -0.04 | +2.19 | +0.64 | +0.4 | +0.33 | +2.19 | +4.35 | +0.19 | -3.01 | +0.13 | +0.86 | +4.39 | +0.69 | -2.26 | -9.28 | +5.39 | +3.29 | +1.49 |

| Method | cow | cup | curtain | dog | door | fence | floor | flower | food | grass | ground | horse | kboard | light | mbike | mountain | mouse | person | plate | pform | mIoU |
|---|---|---|---|---|---|---|---|---|---|---|---|---|---|---|---|---|---|---|---|---|---|
| Zero-Shot | 90.19 | 36.79 | 56.02 | 89.01 | 25.00 | 40.63 | 62.16 | 33.63 | 39.15 | 79.8 | 22.86 | 89.78 | 82.31 | 46.2 | 87.28 | 58.64 | 45.3 | 87.31 | 6.74 | 32.95 | 55.52 |
| Ours | 90.61 | 36.31 | 55.54 | 89.67 | 25.01 | 40.84 | 66.12 | 21.90 | 44.04 | 81.82 | 40.35 | 90.37 | 81.14 | 46.93 | 88.01 | 59.39 | 50.15 | 87.37 | 8.45 | 40.63 | 57.01 |
| | +0.42 | -0.48 | -0.48 | +0.66 | +0.01 | +0.21 | +3.96 | -11.73 | +4.89 | +2.02 | +17.49 | +0.59 | -1.17 | +0.73 | +0.73 | +0.75 | +4.85 | +0.06 | +1.71 | +7.68 | +1.49 |

| Method | plant | road | rock | sheep | shelves | s.walk | sign | sky | snow | sofa | table | track | train | truck | tv | wall | water | window | wood | mIoU |
|---|---|---|---|---|---|---|---|---|---|---|---|---|---|---|---|---|---|---|---|---|
| Zero-Shot | 59.91 | 49.07 | 47.22 | 90.34 | 28.69 | 13.55 | 29.21 | 93.76 | 74.71 | 45.84 | 57.81 | 33.74 | 83.43 | 78.89 | 11.33 | 74.55 | 62.84 | 89.36 | 39.55 | 55.52 |
| Ours | 61.09 | 50.92 | 51.61 | 90.57 | 32.13 | 19.28 | 31.77 | 93.97 | 74.11 | 40.04 | 56.86 | 52.04 | 87.90 | 78.46 | 13.91 | 77.34 | 61.91 | 90.72 | 37.91 | 57.01 |
| | +1.18 | +1.85 | +4.39 | +0.23 | +3.44 | +5.73 | +2.56 | +0.21 | -0.6 | -5.8 | -0.95 | +18.3 | +4.47 | -0.43 | +2.58 | +2.79 | -0.93 | +1.36 | -1.64 | +6.64 | +1.49 |

**Table 4: Per-class results on PASCAL-Context 59.** We compare our proposed method to the Zero-Shot predictions by CAT-Seg.

| Method | chand. | bus | barrel | bridge | poster | schair | wpane | tank | palm | howel | k.island | tower | fountain | ball | boat | build. | rug | shelf | pot | skyscraper | mIoU |
|---|---|---|---|---|---|---|---|---|---|---|---|---|---|---|---|---|---|---|---|---|---|
| Zero-Shot | 33.83 | 71.00 | 6.36 | 34.93 | 7.79 | 4.34 | 42.19 | 1.43 | 25.81 | 0.14 | 8.29 | 21.60 | 36.23 | 54.75 | 65.93 | 60.58 | 46.47 | 27.10 | 24.39 | 25.75 | 26.88 |
| Ours | 51.37 | 85.17 | 15.75 | 43.32 | 14.76 | 10.85 | 48.41 | 7.60 | 31.15 | 5.47 | 3.53 | 16.14 | 29.90 | 47.41 | 57.94 | 52.24 | 38.06 | 17.17 | 13.82 | 14.91 | 27.39 |
| | +17.54 | +14.17 | +9.39 | +8.39 | +6.97 | +6.51 | +6.22 | +6.17 | +5.34 | +5.33 | -4.76 | -5.46 | -6.33 | -7.34 | -7.99 | -8.34 | -8.41 | -9.93 | -10.57 | -10.84 | +0.51 |

**Table 5: Top 10 best and worst per-class results on ADE20k** We compare our proposed method to the Zero-Shot predictions by CAT-Seg.

comes from the ego-vehicle, which is present during both training and evaluation. We thus decide to run an evaluation in which we crop out the bottom 23% of the image, ensuring the ego-vehicle is completely removed from every image. We run this evaluation both before and after the adaptation, and collect results in Table 3. We can notice that the ego-vehicle worsens the performance for the class *car* also for the original Cat-SEG model – since the mIoU already rises from 76.38 (see main paper) to 81.19. Furthermore, by running our adaptation strategy in this setting, our result on this class gets even better, reaching a score of 85.67.

# 3 Class-by-Class Analysis

In the main paper, on PASCAL-Context 59 and ADE20k we only reported the mIoU averaged over all semantic classes for the sake of space. Here, we report more detailed results concerning single classes.

Table 4 collects the mIoU for each single semantic class in PASCAL-Context 59. We can appreciate how the majority of the classes are improved by our method, with only 15 out of the total 59 classes showing some drops.

Table 5 shows the results concerning the 20 classes for which we had the highest increase/drop in mIoU on ADE20k. We can appreciate how classes such as *bus* and *chand.* are largely improved by more than 10%. As a counterpoint, others such as *pot* and *skyscrapers* show a similar drop in performance.

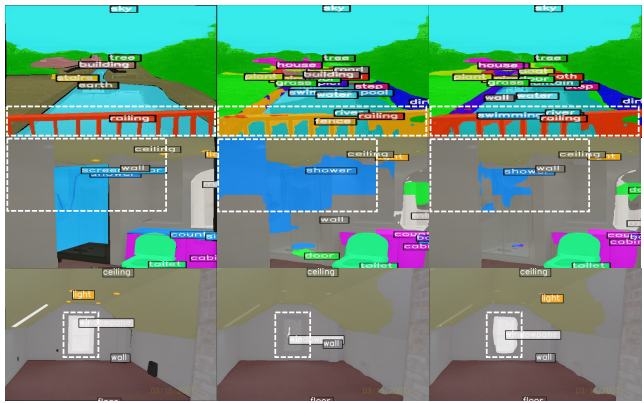

Figure 2: **Qualitative results on ADE20K dataset.** Starting from the *left*, we show respectively the ground truth segmentation map, the zero-shot CAT-Seg prediction, and the result after our adaptation process.

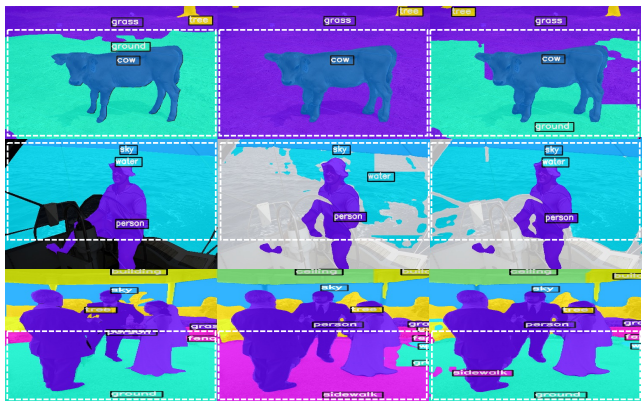

Figure 3: **Qualitative results on PC-59 dataset.** Starting from the *left*, we show respectively the ground truth segmentation map, the zero-shot CAT-Seg prediction, and the result after our adaptation process.

## 4 Qualitative Results

We conclude with some additional qualitative results.

Figure 2 shows some examples from the ADE20k dataset over three rows, respectively overlayed, from left to right, with ground-truth semantic masks, semantic labels predicted by the original CAT-Seg model, or by the one adapted with our strategy. In the first row, we can appreciate how the adaptation process allows CAT-Seg to fully recover the railing semantic class, which was almost lost by the zero-shot model. In the second example, the original CAT-Seg model wrongly labelled a large portion of the scene as part of the *shower* class, whereas it can limit such area to the actual shower visible in the image. Finally, in the third example, we can appreciate how CAT-Seg can properly label even very small objects in the scene – such as the *light* class appearing on the roof, which was completely ignored by the original CAT-Seg weights.

Figure 2 reports three examples from PASCAL-Context 59. Here, in particular, we can appreciate how the adaptation process allows for recovering large regions of the background, being incorrectly labeled as *grass* (top) or *sidewalk* (bottom) by the zero-shot model.


}

\end{document}